\documentclass{article}

\usepackage{arxiv}
\usepackage{xcolor}

\usepackage[utf8]{inputenc} 
\usepackage[T1]{fontenc}    
\usepackage{hyperref}       
\usepackage{url}            
\usepackage{booktabs}       
\usepackage{amsfonts}       
\usepackage{nicefrac}       
\usepackage{microtype}      
\usepackage{lipsum}
\usepackage{graphicx}
\usepackage{titlesec}
\usepackage{amsmath}
\usepackage[euler]{textgreek}
\graphicspath{ {./images/} }
\usepackage[noadjust]{cite}

\title{An FNet based Auto Encoder for Long Sequence News Story Generation}

\author{ \href{https://orcid.org/0000-0002-4966-0494}{\includegraphics[scale=0.06]{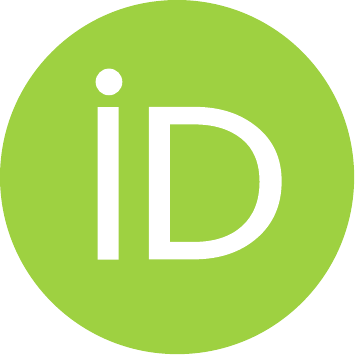}\hspace{1mm}Paul K. Mandal}* \\
  Department of Computer Science\\
  University of Texas at Austin\\
  Austin, TX 78712 USA \\
  \texttt{mandal(at)utexas.edu} \\
  \\
  *Corresponding author\\
   \And
 Rakeshkumar Mahto \\
  Department of Electrical and Computer Engineering\\
  California State University Fullerton\\
  Fullerton, CA 92831 USA \\
  \texttt{ramahto(at)fullerton.edu} \\
}

\begin{document}
\maketitle
\begin{abstract}
 In this paper, we design an auto encoder based off of Google’s FNet Architecture in order to generate text from a subset of news stories contained in Google’s C4 dataset. We discuss previous attempts and methods to generate text from autoencoders and non LLM Models. FNET poses multiple advantages to BERT based encoders in the realm of efficiency which train 80\% faster on GPUs and 70\% faster on TPUs. We then compare outputs of how this autencoder perfroms on different epochs. Finally, we analyze what outputs the encoder produces with different seed texts.
\end{abstract}


\maketitle

\section{Introduction}
Text generation has been a longstanding area of interest within Artificial Intelligence. A non neural network approach proposed by McKeown involved creating a script for different items in a QA system and then using a database that would look up key parameters inquired about a certain system and insert them into the script \cite{b1}. Fortunately, more modern approaches using Machine Learning and Neural Networks are much less tedious. In this paper, we design an autoencoder based off of the FNet architecture proposed by Google \cite{b2}. FNet has significant efficiency advantages over LSTM and BERT based approaches, training 80\% faster on GPUs and 70\% faster on TPUs. In this paper, we propose an FNET based architecture trained on 1,000,000 news stories that is a subset of the C4 dataset \cite{b3}.

\subsection{Previous Research}
In the past 3 years, major developments have been made in the field of sensible text generation by the use of Large Language Models (LLMs) such as GPT NeoX \cite{b19}. However, in order to compose LLMs, many conventional layers of conventional networks such as CNNs and LSTMs are required. In this paper, we discuss the design of an Autoencoder based off of FNET. Additional work will be done in a subsequent paper on implementing a Variational Auto Encoder (VAE) using the architecture outlined here. Both would be crucial to integrate FNet Layers in an LLM model.

Fake News has been a major issue over the past 5 years. Our motivation for using news headlines is to see whether a primitive neural network could generate sensible news like text. In practice, it would be much more effective to train an LLM in order to achieve this purpose. However, development and analysis in the performance of more primitive layers are still crucial to building more effective models.

\subsection{Dataset}
For the development of our neural network, we decided to train on the NewsLike subset of the Colossal, Cleaned Common Craw (C4) dataset collected by huggingface. The "clean" C4 dataset on huggingface is 305 GB. The NewsLike subset of C4 is 15 GB. For our paper, we restricted the training of our neural network to 1,000,000 news stories due to hardware and memory limitations.

\section{Background}
Significant advancements have been made in the field of Natural Language processing by using neural networks. Problems that were thought to have no or extremely difficult solutions have been solved with deep learning. In order to understand the architecture proposed in this paper, a review of the following concepts is necessary.

\subsection{The Perceptron}
Perceptrons work by loosely imitating the way that neurons function in the brian \cite{b5}. While a conventional neuron works by receiving electromagnetic shocks through it’s dendritic tree and determines whether to sent an electric shock through it’s axon by the balance of certain chemicals through a neuron, a perceptron is a loose mathematical of this \cite{b4}. Given an input vector x and a weight vector θ, a neuron or perceptron can be modeled as follows:

We define the sigmoid activation function to be,
\begin{equation}\sigma(z)=\frac{1}{1+e^{-z}}.\label{sigmoid}\end{equation}

The optimal weights, \texttheta, are trained by minimizing the cost function \begin{equation}J(\theta)= - \frac{1}{m} \sum_{i=1}^{m} Cost(\theta^{T}x^{i},y^{i})\label{train}\end{equation}

Where $i$ refers to the $i^{th}$ element and $Cost(\theta^{T}x,y)$ is defined as,

\begin{equation}
Cost(\theta^{T}x,y)=
    \begin{cases}
        -log(\sigma(\theta^{T}x)) & \text{if } y = 1\\
        -log(\sigma(1 - \theta^{T}x)) & \text{if } y = 0
    \end{cases}
\end{equation}

Which we can elegantly and intuitively rewrite as
\begin{equation}
Cost(\theta^{T}x,y)= -ylog(\sigma(\theta^{T}x)) \\
- (1 - y)log(\sigma(1 - \theta^{T}x))
\end{equation}

\begin{figure}
\center
\includegraphics[width=.5\columnwidth]{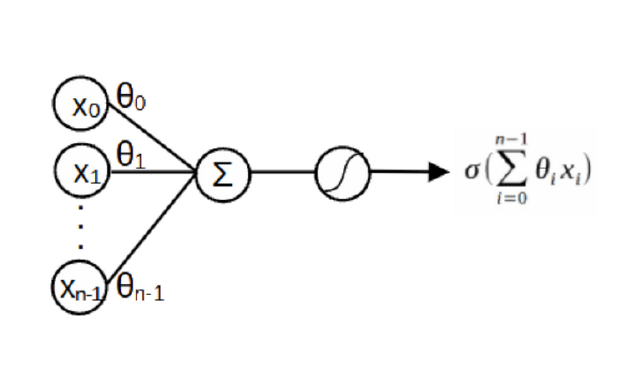}
\caption{A simple neuron classifier with a sigmoid activation function.}\label{fig1}
\end{figure}

\subsection{Feed Forward Neural Networks and the Multi Layer Perceptron}
A feed forward neural network is defined when two layers of neurons form a fully connected graph. Neural networks are able to extract higher level features that assist with classification and regression problems \cite{b6}. Neural networks are trained using backpropagation. Discussion of backpropagation is beyond the scope of this paper, but essentially it is a form of gradient descent where the chain rule is applied to account for each layer of the neural network \cite{b7}.

\begin{figure}
\center
\includegraphics[width=.5\columnwidth]{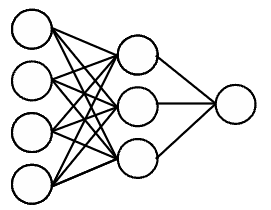}
\caption{A feed forward neural network.}\label{fig2}
\end{figure}

\subsection{Convolutional Neural Networks}

Convolution is a process that is often used for signal processing. For a 1-dimensional discrete sequence, we define convolution to be,
\begin{equation}(f * g)[n] = 
    \sum_{k = -\infty}^{\infty} f[k]g[n - k]
\end{equation}

Where f and g are two discrete functions \cite{b9}.

Convolution is often used in neural networks for object recognition and sequence processing \cite{b10}. The convolution’s property invariance under translation is a useful feature that allows it to recognize objects regardless of where it is in the image. In the case of neural networks, the sequence and filter lengths are finite.

\begin{figure}
\center
\includegraphics[width=.4\columnwidth]{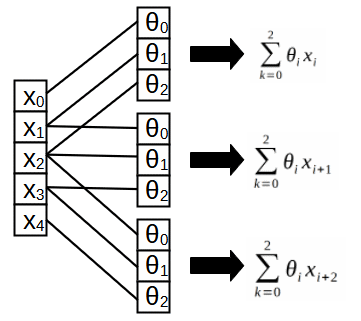}
\caption{An example of 1D Convolution with three input weights applied a sequence of 5 inputs.}\label{fig3}
\end{figure}

\subsection{Embedding Layers}

An Embedding layer is a representational layer that stores words as a vector of real values. These vectors are essentially machine representations of meanings. These vector encodings can allow neural networks to encode much more meaningful representations of words which allow for more accurate performance \cite{b11}. Embeddings can either be learned or alternatively, pretrained embedding layers can be used GloVe \cite{b13}.

\begin{figure}
\center
\includegraphics[width=.4\columnwidth]{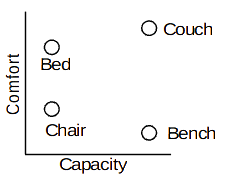}
\caption{An example of how an embedding could conceptualize furniture by using comfort and capacity as parameters.}\label{fig4}
\end{figure}

\subsection{Auto Encoders}
The auto encoder is the main method used in deep learning to generate text. An auto encoder is a neural network that attempts to produce the same output as it’s inputs \cite{b12}. The output of the autoencoder itself isn’t what’s useful; it’s the features that the auto encoder learns useful features due to the reduction in dimensionality from the input to the hidden layer in the encoder part of the model. The decoder portion of the model then expands the model back to the same dimension as the original input. Auto encoders were proven to be useful when a paper written by Hinton and Salakhutdinov demonstrated that a neural network that reduced the dimension to 30 was able to reconstruct more information than principal component analysis algorithm that attempted to compress the same data to the same dimensionality \cite{b14}.

\begin{figure}
\center
\includegraphics[width=.4\columnwidth]{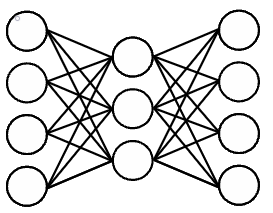}
\caption{A feed-forward autoencoder with 4 input neurons, a 3 neuron hidden layer, and 4 output neurons.}\label{fig5}
\end{figure}

\subsection{Fourier Transforms}
Fourier transforms have a wide variety of uses but are often used in signal processing \cite{b15}. A fourier transform in essence is a conversion of a function from the time domain to the frequency domain. The general fourier transform is defined as,

\begin{equation}F(\omega) = 
    \int_{-\infty}^{\infty} e^{-i\omega t}f(t) dt
\end{equation}

A fourier series is a series of sinusoidal waves. Any periodic signal can be represented as a series of sinusoidal waves multiplied by it’s fourier coefficient. 
\begin{equation}f(t) = 
    \sum_{n = 0}^{N - 1} c_n e^{-i2\pi nt}
\end{equation}

Where each fourier coefficient, $c_n$, can be solved by computing 
\begin{equation}c_n = 
    \frac{1}{L} \int_{-L/2}^{L/2} e^{-i2\pi \frac{n}{L} t} f(t) dt
\end{equation}

A fourier transform provides a few advantages in the context of a neural network. It allows us to perform convolution while still leveraging the advantages of a time encoding that an LSTM brings. However, LSTMs are more computationally expensive than performing convolution.

\section{Model Architecture}
As previously mentioned, the architecture of our model is adapted from FNet \cite{b2}. There are a few modifications on our network from FNet. FNet normally performs two fourier transforms, one along the sequence dimension and one along the hidden dimension. Because the Fourier Transform is commutative, it does not matter which transform is applied first. For our neural network, we only perform it along the sequence dimension. 

We then append the output from our embedding dimension and our fourier transform, normalize it,  and put that through a feed forward neural network. We then again normalize the data, put it through a dense layer, and then connect it to the decoder which expands the dimensionality of the data back to the original dimension. 

This architecture was programmed in Keras and Tensorflow \cite{b16} \cite{b17}. Some of our code for FNet was borrowed from the Keras website \cite{b18}. Our model had 4,977,808 parameters and was trained on 800,000 news stories and validated over 100,000. For our parameters, we used a dictionary size of 10000 and stemmed news stories to 150 words. Our embedding dimension was 128, our latent dimension was 64, our number of heads were 8, and we used a batch size of 64.

\begin{figure}
\center
\includegraphics[width=.5\columnwidth]{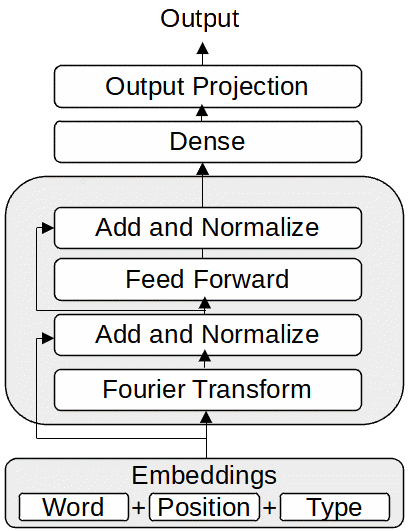}
\caption{A block diagram of the FNet Architecture}\label{fig6}
\end{figure}

\section{Results}

We trained this neural network on a computer with a Xeon W-10855M (a 12 core, 24 thread processor), a NVIDIA 2080S, and 64 GB of RAM. It took our neural network about 25 hours to train on 50 epochs. With this model, we were able to achieve a 96.36\% accuracy. Although it was asymptotic, loss continued to lower into the 50th epoch indicating that the data was not overfitting. Prima facie, this may appear that this is too close to a one to one correspondence for the model to learn useful features, but our generated text from different seed text inputs indicates otherwise.

A JSON file containing our results is available at \url{https://github.com/PaulKMandal/FNET_News_Headlines}. As can be seen from our results, much of the text that our model produces is nonsensical. As previously mentioned, the main goal of this paper wasn't to produce sensible text as an LLM would be best suited for that task. Here, we lay the foundation for eventually building an FNET VAE which could be used in much larger models.

\section{Future Work}

More insightful models could be created by further work in hyperparameter tuning. Using glove embeddings could also prove fruitful for more general applications of this autoencoder. However, we plan to pursue further work in this model in two specific areas. 

The first whether it is possible to create an FNet Variational Autoencoder. A variational autoencoder allows someone to generate an output based off of a continuous representation of data in the intermediate layer instead of using some seed input (for example entering a seed text or seed image into the input layer to attempt to produce an output). Although conventional autoencoders are useful for certain problems such as denoising, designing a variational autoencoder generates much more insightful outputs since generating text by using the full range of the intermediate layers continuous domain is much less constrained as opposed to the limited amount of permutations that can be used for seed text.

Secondly, we also plan on implementing an FNET based model for a generative adversarial network. GANs can provide much more convincing outputs than their pure autoencoder counterparts. The main challenge with generative adversarial networks is tuning hyperparameters in both the generator and discriminator so that they can train.

\end{document}